\documentclass[11pt,a4paper]{article}
\usepackage[hyperref]{emnlp2020}
\usepackage{times}
\usepackage{latexsym}
\usepackage{graphicx}
\usepackage{float}
\usepackage{amsfonts}
\usepackage{amsmath}
\usepackage{xcolor}

\usepackage{microtype}

\aclfinalcopy 

\newcommand\blfootnote[1]{%
  \begingroup
  \renewcommand\thefootnote{}\footnote{#1}%
  \addtocounter{footnote}{-1}%
  \endgroup
}
\newcommand{\specialcell}[2][c]{%
  \begin{tabular}[#1]{@{}c@{}}#2\end{tabular}}

\title{Sentiment Analysis for Reinforcement Learning}

\author{Ameet Deshpande\textsuperscript{*}, Eve Fleisig\textsuperscript{*}  \\
        Princeton University \\
  \texttt{\{asd, efleisig\}@cs.princeton.edu} \\

}

\date{}

\begin{document}
\maketitle
\blfootnote{\textsuperscript{*}Equal contribution}

\begin{abstract}
While reinforcement learning (RL) has been successful in natural language processing (NLP) domains such as dialogue generation and text-based games, it typically faces the problem of sparse rewards that leads to slow or no convergence. Traditional methods that use text descriptions to extract only a state representation ignore the feedback inherently present in them. In text-based games, for example, descriptions like ``\textit{Good Job! You ate the food}'' indicate progress, and descriptions like ``\textit{You entered a new room}'' indicate exploration. Positive and negative cues like these can be converted to rewards through sentiment analysis. This technique converts the sparse reward problem into a dense one, which is easier to solve. Furthermore, this can enable reinforcement learning without rewards, in which the agent learns entirely from these intrinsic sentiment rewards. This framework is similar to intrinsic motivation, where the environment does not necessarily provide the rewards, but the agent analyzes and realizes them by itself. We find that providing dense rewards in text-based games using sentiment analysis improves performance under some conditions.
\end{abstract}


\section{Introduction}


Reinforcement learning has shown great success in environments with large state spaces. Using neural networks to capture state representations has allowed end-to-end training of agents on domains like Atari \cite{mnih2015human} and Go \cite{silver2016mastering}. It is natural to emulate this success in text domains, especially given that the state space in language-based tasks is combinatorially large. A sentence of length $l$ with allowed vocabulary $|V|$ has $O(|V|^l)$ possible states, and tabular methods like $Q-$learning \cite{watkins1992q} will fail unless coupled with powerful function approximators like neural networks.\\

\begin{figure}[ht]
\begin{center}
\centerline{\includegraphics[width=\columnwidth]{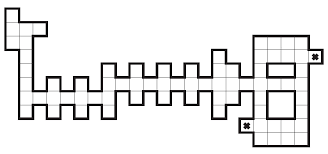}}
\caption{A sparse reward setting where the agent gets a reward signal only in two states.}
\label{fig:sparse}
\end{center}
\vskip -0.2in
\end{figure}

\begin{figure}[ht]
\begin{center}
\centerline{\includegraphics[width=\columnwidth]{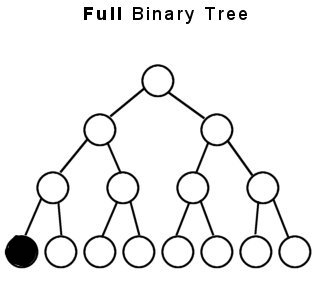}}
\caption{Full binary tree with single reward}
\label{fig:binary}
\end{center}
\vskip -0.2in
\end{figure}

While the current state of RL has multiple challenges, sparse rewards are one that leads to slow, and sometimes no convergence. Consider an agent learning in an environment with a large state space, with only a few states leading to a reward (Figure \ref{fig:sparse}). An agent starting on the far left must take a large number of actions before encountering a reward. In turn, this sparse feedback results in a very noisy gradient for training the neural network. In an extreme scenario, as in Figure \ref{fig:binary}, an agent might have to take an exponential number of actions to reach a single leaf that has a reward.

Some early work, such as reward shaping \cite{ng1999policy}, attempted to solve the sparse reward problem by introducing dense rewards based on heuristics, e.g., how close the agent is to the goal. However, these require complex design choices that might result in unexpected behavior from the agents.\\

Sparse rewards are common because they are the most straightforward way to specify how a task needs to be solved. If a robot is expected to pour water from a jug into a glass, the simplest way is to give a reward of $1$ if it fills the glass, and $0$ otherwise. This type of reward design is common in text-based games, in which the agent is rewarded upon reaching the goal state, and task-oriented dialogue, in which the agent is rewarded based on the successful completion of the task.\\

For this study, we examine text-based games and find that providing dense rewards with the help of sentiment analysis improves performance under some conditions.

\section{Motivation}

Language understanding for reinforcement learning faces challenges that include partial observability, large action spaces, balancing exploration and exploitation, and sparse rewards. Environments may give the same feedback for different commands, and information may not be apparent in an observation (e.g., whether a door is locked). Moreover, because the environment may contain many possible states or actions, reinforcement learning methods must approximate solutions to determine what commands are valid and useful. Agents must also determine what game cues indicate directions to explore and balance deliberately exploring the game space with exploiting local rewards.

This work addresses the underlying problem of long-term credit assignment: how to determine what actions are responsible for obtaining a reward. Because rewards may be sparse, agents must be able to generate a sequence of actions before any change in the environment or a reward signal (e.g., "You enter a new room" or "Your score has gone up by one point"). In most text-based games, agents must be able to perform on the order of 10-20 steps without a reward when following an optimal trajectory \cite{textworld}. By contrast, the presence of dense rewards allows more immediate feedback and aids in identifying which steps lead to a reward.

\begin{figure}[ht]
\begin{center}
\centerline{\includegraphics[width=\columnwidth]{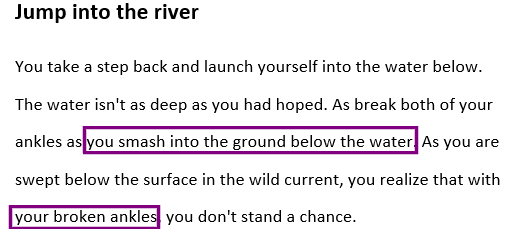}}
\caption{Fatal Island: Extracting negative sentiment}
\label{fatal}
\end{center}
\vskip -0.2in
\end{figure}

\begin{figure}[ht]
\begin{center}
\centerline{\includegraphics[width=\columnwidth]{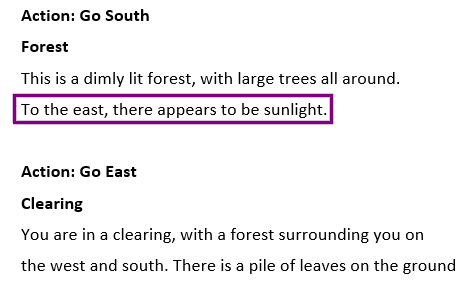}}
\caption{Zork: Exploration cue}
\label{zork}
\end{center}
\end{figure}

However, text-based games do contain sentiment-laden descriptions associated with eventual rewards and cues to explore certain game areas. Figure \ref{fatal} gives an example of negative sentiment in the game Fatal Island: negative statements such as ``you smash into the ground" and ``broken ankles" indicate that this path is not the one to take, although no reward is associated with the description. In the game Zork, exploratory cues such as ``To the east, there appears to be sunlight" in an otherwise nondescript forest suggest that the agent should explore in that direction, even though no reward is immediately given for doing so (Figure \ref{zork}). Thus, the ability to convert these descriptions into rewards provides the capacity to train with dense rewards, increasing an agent's ability to identify important states to explore or actions to take that lead to successful trajectories.

\section{Related Work}
\subsection{Language-Aided Reinforcement Learning}

Recent work has begun incorporating information from NLP into the rewards used in reinforcement learning tasks, particularly dialogue generation and text-based games. \citet{DBLP:journals/corr/LiMRGGJ16} trained two reinforcement learning agents to produce less repetitive, more coherent responses using rewards based on seq2seq models of the responses. \citet{DBLP:journals/corr/abs-1708-00133} mapped text descriptions to transitions and rewards in an environment by creating a Q-function for reinforcement learning conditioned on the descriptions. Focusing on the problem of finding intermediate rewards, \citet{DBLP:journals/corr/abs-1903-02020} incorporated GloVe vectors into their reward function, significantly improving performance on text-based games. However, these reinforcement learning approaches have not yet incorporated the ability to fine-tune BERT \citep{DBLP:journals/corr/abs-1810-04805} to set the rewards given to the user output on a given turn (for dialogue) or the appearance of a new text description (for text-based games).\\

Some studies have used natural language as a decoy for reward functions \cite{macglashan2015grounding, williams2018learning}, but the text used in them must explicitly indicate what the goal state is, and an object-oriented MDP \cite{diuk2008object} must then be used to extract the goal state. These restrictions severely limit the use cases of the method. Others have examined using text descriptions from a game manual to extract information relevant to determining rewards and promising states during gameplay, though without analyzing text during gameplay itself \cite{branavan-etal-2012-learning}.
\citet{krening2017learning} acknowledge the dearth of methods that use text explanations and advice, but instead propose a separate policy for each object, which hinders scaling up this method.

\subsection{Sparse Reward RL}

Alleviating the sparse reward problem has received much attention since the inception of RL. We note some research that is orthogonal to our proposed method and can thus be used in tandem. Reward shaping gives auxiliary rewards to an agent based on how ``far" in the state space the agent is from the goal \cite{ng1999policy}. For example, for an agent that must navigate to a goal state on a 2-dimensional grid, the auxiliary reward can be inversely proportional to the distance from the goal state on the grid. This method requires the additional restriction that the reward function be a potential-based energy function to avoid positive reward loops.\\

Hindsight Experience Replay uses the novel idea that both successful and unsuccessful trajectories can be used for training if the policy learned is conditional on the goal \cite{andrychowicz2017hindsight}. For example, if an agent tries to kick a ball straight ahead and ends up kicking it to the left, it can use that trajectory by assuming it actually wanted to kick it to the left, hence increasing the number of trajectories on which it receives a positive reward.\\

Other work, such as \citet{agarwal2019learning}, makes use of a meta-framework to provide auxiliary rewards that supplement the environment's rewards. Our method resembles this in a limited setting.


\subsection{Recent Text-Based Game Models}

More recent approaches to text-based games have examined other methods in which natural language can inform RL agents. \citet{ammanabrolu2018playing} represented the game state as a knowledge graph learned during game exploration. This method allowed the agent's action space to be reduced to a question-answering task, pruning options to allow for more efficient exploration. Work on affordance extraction--determining the set of behaviors enabled by a particular game state--has advanced by using word embeddings to create a common knowledge database. The database can then be queried by the RL agent; affordance-based action improved performance in most situations \cite{fulda2017can}. Although some of these efforts have goals that eventually diverge from our research, they share the need for denser rewards, an overarching issue in the success of linguistically informed approaches to text-based games.

\section{Methodology}

\begin{figure}[ht]
\vskip 0.2in
\begin{center}
\centerline{\includegraphics[width=\columnwidth]{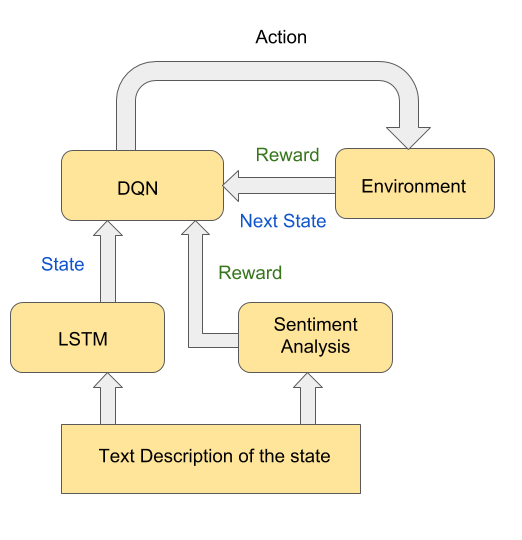}}
\caption{Model}
\label{model}
\end{center}
\vskip -0.2in
\end{figure}

Figure \ref{model} depicts our methodology. Given a text description representing a state, we use an LSTM \cite{hochreiter1997long} to encode the state representation and feed that into the Deep-Q Network (DQN) \cite{mnih2015human}. We then use the reward given by the environment and supplement it with a sentiment-based reward extracted from the state representation. Thus, even if the reward from the environment is $0$, the sentiment-based reward allows it to potentially learn faster.


For the sentiment analyzer, we assume we have access to positive and negative trajectories in the environment. Negative trajectories are generated by running a random agent on the environment because its success rate is low. To generate positive trajectories, we use the walkthroughs associated with the games, which follow the correct set of actions to reach the goal.\\

Mathematically, the new reward sent to the agent is:
\[  r_{total} = r_{env} + scale\times polarity     \]
where the sentiment analyzer is assumed to be a binary classifier that outputs a continuous score between $[-1,1]$, called polarity.

\section{Model, Datasets and Environment}

\subsection{Reinforcement Learning Model}
\begin{figure}[ht]
\vskip 0.2in
\begin{center}
\centerline{\includegraphics[width=\columnwidth]{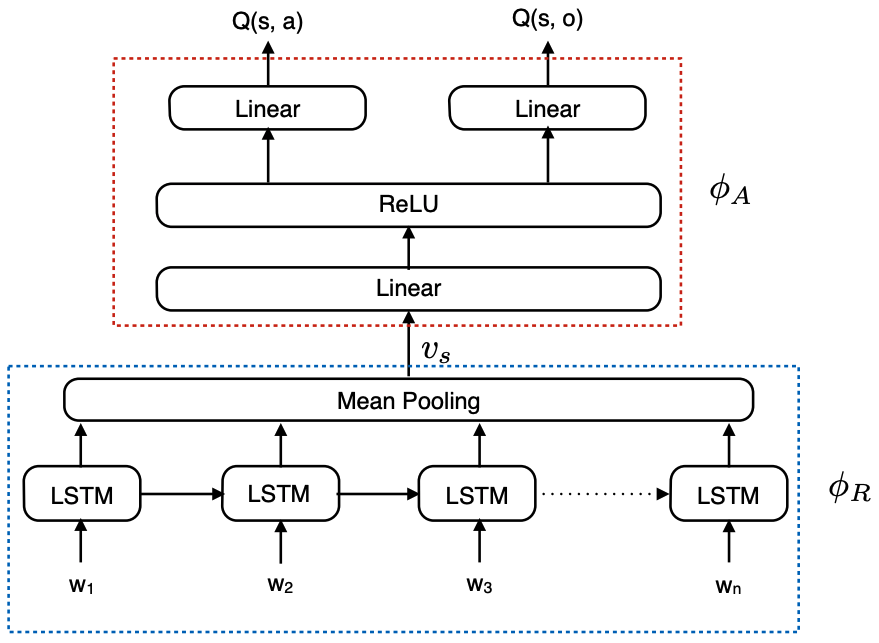}}
\caption{Model}
\label{fig:lstmdqn}
\end{center}
\vskip -0.2in
\end{figure}

We explain the different models we use in our method. The agent follows the LSTM-DQN in \citet{narasimhan-barzilay-2015-machine} closely. The model uses the standard $Q-$learning equation for training.
\[ Q_{i+1}(s,a) = \mathbb{E}[r+\gamma \max_{a'} Q_i(s',a') | s,a]  \]
The LSTM receives the words as input and produces a state representation, which is the average of all the final outputs of the LSTM. This is then fed to a 2-layer neural network that calculates the $Q$ scores. Unlike \citet{narasimhan-barzilay-2015-machine}, we do not use different $Q$ scores for the action and the object. The model also uses prioritized experience replay \cite{schaul2015prioritized} so that successful trajectories can be replayed more often than unsuccessful ones. This can be viewed as another way of alleviating the sparse rewards problem. The full model is shown in Figure \ref{fig:lstmdqn}. Although the LSTM-DQN model is not state-of-the-art, we believe that because our modifications change the environment, the modifications can be applied to any method in future work.\\

\subsection{Environment}
The models were tested on the TextWorld learning environment for text-based games. TextWorld problems are centered on cooking a recipe, requiring agents to determine the necessary ingredients from a recipe book, explore a house to gather the ingredients, and cook the meal in the kitchen with specific tools. The agent must also overcome obstacles, such as locked doors. The obstacles, ingredients, recipes, and environment layouts change between games \cite{textworld}.

We fine-tuned and tested the models to extract sentiment from a dataset of 1,048 TextWorld game trajectories (524 wins and 524 losses). Each model was then tested on the TextWorld environment.

\subsection{Sentiment Models}\label{section:sentiment-models}
We fine-tuned BERT \citep{DBLP:journals/corr/abs-1810-04805} to classify game trajectories as wins or losses based on the sentiment of the text descriptions. The learning rate was varied among $\{1 \times 10^{-6}, 1 \times 10^{-5}, 2 \times 10^{-5}, 1 \times 10^{-4}\}$ and the number of training epochs from 500 to 1500. The final model was trained with a learning rate of $2 \times 10^{-5}$ for 1500 epochs.

In addition, we used a TextBlob model, a naive Bayes classifier trained on the Stanford movie review dataset \cite{maas-etal-2011-learning}. We also trained a naive Bayes classifier on the bag-of-words representations of the text descriptions (smoothed with $\alpha=1.0$), with positive and negative trajectories as positive and negative instances.

\subsection{Training on Auxiliary Data}

\begin{figure}[ht]
\begin{center}
\centerline{\includegraphics[width=\columnwidth]{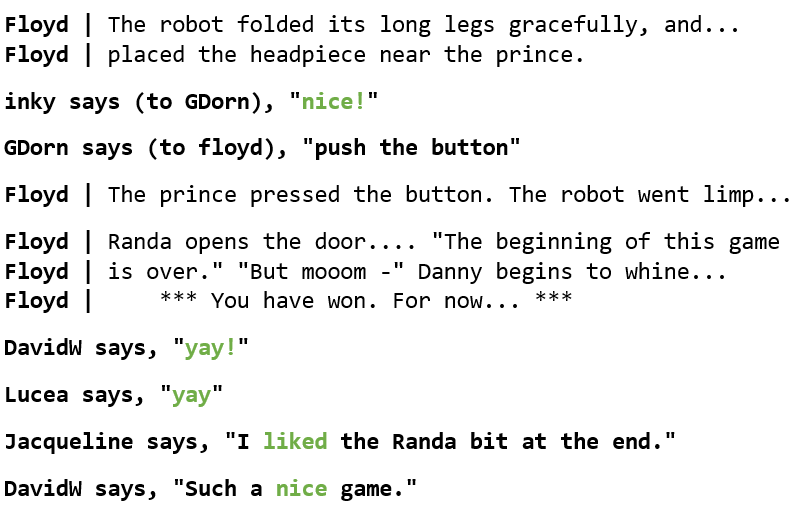}}
\caption{Example of a winning ClubFloyd trajectory with positive user banter.}
\label{fig:banter_ex}
\end{center}
\vskip -0.2in
\end{figure}

As an extension of our work, we examined how sentiment could be extracted from game trajectories beyond examining the text descriptions themselves. Because text produced by human users as they play a game might have higher levels of sentiment than prewritten game text, we also trained on a dataset of games scraped from the interactive fiction website ClubFloyd, which provides transcripts of users playing text-based games that includes messages between players alongside the system output. The players' chat messages include banter that does not change the game state but does indicate the players' reactions to them. For example, player banter contains comments like "nice!" and "yay!" upon performing actions that lead to a winning state (Figure \ref{fig:banter_ex}).

We hypothesized that this banter could provide a better indication of the sentiment associated with a given state. We also hypothesized that the correlation between the degree of positive banter sentiment and the success of a trajectory would increase nearer to the end of a trajectory.

We fine-tuned BERT to predict whether a text description was associated with positive or negative banter. The learning rate was varied among $\{1e-6, 1e-5, 2e-5, 1e-4\}$ and the number of training epochs from 500 to 2000. The final model was trained with a learning rate of $2 \times 10^{-5}$ for 2000 epochs. 

\begin{table}
\centering
\begin{tabular}{lrll}
\hline \textbf{Dataset} & \textbf{F1} & \textbf{Precision} & \textbf{Recall} \\ \hline
ClubFloyd banter & 0.898 & 0.946 & 0.854\\
TextWorld output & 0.751 & 0.839 & 0.680\\
\hline
\end{tabular}
\caption{\label{table:ft_results} Results for ClubFloyd banter and TextWorld system output fine-tuning. }
\end{table}




\section{Analysis}

After fine-tuning the BERT model on the TextWorld trajectories and on the ClubFloyd banter, we examined the models' performance and the correlation between the sentiment scores and the success of the trajectory.

\begin{table*}
\centering
\begin{tabular}{| c | c | c | c | c | c | c |}
\hline 
\textbf{\specialcell{Mean + Sentiment\\for Wins}} & \textbf{\specialcell{Mean + Sentiment\\ for Losses}} & $\sigma$ & \textbf{\specialcell{Spearman\\correlation}} & $p$ & \textbf{\specialcell{Point-biserial\\correlation}} & $p$\\ \hline
0.726 & 0.146 & 0.365 & 0.713 & $<10^{-45}$ & 0.758 & $<10^{-45}$
\\
\hline
\end{tabular}
\caption{\label{table:banter_correl1} Correlation between degree of positive trajectory sentiment and trajectory success on the fine-tuned trajectory model.}
\end{table*}

The model classified game trajectories as wins or losses based on the sentiment of the text descriptions with $F_1=0.898$ (Table \ref{table:ft_results}). We examined the correlation between the sentiment scores on the model trained on TextWorld trajectories and the success of the trajectory. There was a significant, comparatively strong correlation ($r_s=0.713$) between the mean positive sentiment in a trajectory and its success.

\begin{table*}
\centering
\begin{tabular}{| c | c | c | c | c |}
\hline \textbf{\specialcell{Number of turns\\before end}} & \textbf{\specialcell{Mean + Sentiment Score for\\Win Trajectories}} & \textbf{\specialcell{Mean + Sentiment Score for\\Loss Trajectories}} & \textbf{Difference} & $\mathbf{\sigma}$ \\ \hline
5 & 0.722 & 0.703 & 0.019 & 0.158\\
10 & 0.726 & 0.697 & 0.029 & 0.156\\
15 & 0.730 & 0.695 & 0.034 & 0.155\\
20 & 0.731 & 0.696 & 0.035 & 0.153\\
35 & 0.732 & 0.696 & 0.036 & 0.153\\
50 & 0.732 & 0.696 & 0.036 & 0.152\\
100 & 0.733 & 0.693 & 0.039 & 0.154\\
\hline
\end{tabular}
\caption{\label{table:banter_correl2} Mean positive sentiment for banter in last $k$ rounds of ClubFloyd win or loss trajectory.}
\end{table*}

\begin{figure}
\begin{center}
\centerline{\includegraphics[width=\columnwidth]{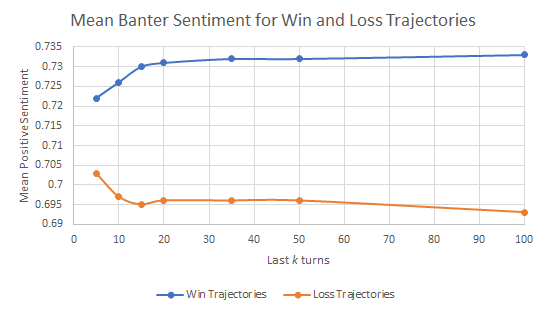}}
\caption{Mean positive sentiment for ClubFloyd banter in last $k$ turns of a win or loss trajectory.}
\label{fig:last-k}
\end{center}
\vskip -0.2in
\end{figure}

\begin{table*}
\centering
\begin{tabular}{| c | c | c | c | c |}
\hline \textbf{\specialcell{Number of turns\\before end}} & \textbf{\specialcell{Spearman\\ correlation}} & $p$ & \textbf{\specialcell{Point-biserial\\ correlation}} & $p$\\ \hline
5 & 0.047 & 0.002 & 0.045 & 0.004\\
10 & 0.061 & $1.41 \times 10^{-7}$ & 0.070 & $9.15 \times 10^{-10}$\\
15 & 0.071 & $3.73 \times 10^{-13}$ & 0.087 & $6.32 \times 10^{-19}$\\
20 & 0.077 & $9.27 \times 10^{-19}$ & 0.089 & $3.50 \times 10^{-29}$\\
35 & 0.083 & $2.65 \times 10^{-32}$ & 0.096 & $7.10 \times 10^{-43}$\\
50 & 0.081 & $3.924 \times 10^{-40}$ & 0.095 & $1.16 \times 10^{-53}$\\
100 & 0.093 & $2.94 \times 10^{-81}$ & 0.108 & $2.12 \times 10^{-109}$\\
\hline
\end{tabular}
\caption{\label{table:banter_correl} Correlation between degree of positive/negative banter sentiment and trajectory success for last $k$ turns of a trajectory.}
\end{table*}

The model also classified text descriptions as corresponding to negative or positive banter with $F_1=0.751$. We found, as hypothesized, that the banter in winning trajectories would have a higher degree of positive sentiment (Figure \ref{fig:last-k}). However, this difference remains small, resulting in a low but significant correlation between trajectory success and mean positive sentiment (Table \ref{table:banter_correl}). We also found that this correlation strengthens as the number of turns examined preceding a win or loss is increased.

\section{Results and Discussion}

We present the results after trying multiple sentiment analysis models. The environments that we train on typically have a reward for reaching the goal state, and some intermediate rewards for reaching other crucial states. We evaluate each model in two settings, one in which these intermediate rewards are present ([Model]), and one in which only the final reward is present ([Model-Zero]), where $[.]$ denotes a placeholder. We use the aforementioned equation to supply the total reward:

\[  r_{total} = r_{env} + scale\times polarity     \]

where $scale$ is a hyperparameter of the model, tuned for several values between $[0,1]$, and we use $0.1$ for the final value. This avoids the issue that larger values of $scale$ could allow the agent to accrue enough auxiliary reward that it would not have to reach the goal state at all.

\subsection{Naive Bayes}

We use two naive Bayes models (see Section \ref{section:sentiment-models}): the TextBlob model, which is trained on a general corpus of movie reviews, and a Naive Bayes model that we trained on the positive and negative trajectories that we collected. The four versions are \texttt{TextBlob, TextBlob-Zero, Naive-Bayes, Naive-Bayes-Zero}.

\subsection{BERT}

For the BERT model, we use a total of $3$ variants. One is fine-tuned on the Stanford movie review dataset \cite{maas-etal-2011-learning}, one is fine-tuned on the positive and negative trajectories, and the last is fine-tuned on banter. These are called \texttt{BERT-Stanford, BERT-Stanford-Zero, BERT-Traj, BERT-Traj-Zero, BERT-Banter}.\\

Another variant we try is to use the polarity of the state representation only when the model is confident about it, which we call the \textit{threshold version}. This is motivated by the fact that in binary classification, the model is most confused about data points that are close to the boundary. Using these noisy polarity values might be detrimental to the model. Thus, we use the following formula to calculate the polarity: 
\begin{align*}
    \begin{split}
         polarity = \begin{cases} polarity &\mbox{if } polarity > 0.7 \\
         polarity &\mbox{if } polarity < -0.7 \\
         0 &otherwise \\
        \end{cases}
    \end{split}
\end{align*}
Since the threshold version performs better, unless otherwise specified, the threshold version is used.

The baselines that we use, which are models that do not receive any auxiliary rewards, are called \texttt{Vanilla, Vanilla-Zero}.

\subsection{Results}

We use the scores accrued in the training games as the evaluation metric. While it is unusual to evaluate on train data in NLP, in RL it is standard to try to solve the task as best possible, because it is assumed that at evaluation time the agent is interacting with the same environment it was trained on. This is similar to the metrics used in seminal work \cite{mnih2015human}.

To make the comparisons fair, we report only the rewards received from the environments (both intermediate and final). We do not add the auxiliary rewards that our models receive while reporting the scores because the \texttt{Vanilla} models do not have access to them. Thus, the results directly reflect task success.\\

\begin{figure}[ht]
\vskip 0.2in
\begin{center}
\centerline{\includegraphics[width=\columnwidth]{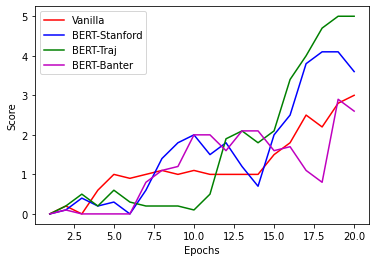}}
\caption{Scores as training progresses}
\label{fig:scores}
\end{center}
\vskip -0.2in
\end{figure}

We train the models for 20 epochs on $10$ games sampled from TextWorld and report the train score curves and the final aggregated curves. For clarity, we report only the best variants in Figure \ref{fig:scores}, and all the models in Table \ref{table:results}. The aggregated score is the sum of scores over all epochs, and the max score represents the maximum score that was achieved over the $20$ epochs. While both metrics are useful, the max score metric is better because a model may reach its peak performance only at the end of the training and therefore have lower aggregated scores. Higher aggregated scores mean lower regret, where regret is the difference between the score of an optimal agent and the score of our agent, aggregated over all the episodes.

\begin{table*}[ht]
\centering
\begin{tabular}{| l | c | c |}
\hline \textbf{\specialcell{Model}} & \textbf{\specialcell{Aggregated}} & \textbf{Max Score}\\ \hline
\texttt{\textbf{Vanilla}} & 24.7 & 3.0\\ \hline
\texttt{\textbf{Vanilla-Zero}} & \textcolor{gray}{0.3} & \textcolor{gray}{0.2}\\ \hline
\texttt{\textbf{TextBlob}} & \textcolor{gray}{0.7} & \textcolor{gray}{0.3}\\ \hline
\texttt{\textbf{TextBlob-Zero}} & \textcolor{gray}{0.0} & \textcolor{gray}{0.0}\\ \hline
\texttt{\textbf{Naive-Bayes}} & \textcolor{gray}{0.8} & \textcolor{gray}{0.3}\\ \hline
\texttt{\textbf{Naive-Bayes-Zero}} & \textcolor{gray}{0.7} & \textcolor{gray}{0.3}\\ \hline
\texttt{\textbf{BERT-Stanford}} & \textbf{32.1} & 4.1\\ \hline
\texttt{\textbf{BERT-Stanford-Zero}} & \textcolor{gray}{0.7} & \textcolor{gray}{0.2}\\ \hline
\texttt{\textbf{BERT-Traj}} & \textbf{33.0} & \textbf{5.0}\\ \hline
\texttt{\textbf{BERT-Traj-Zero}} & \textcolor{gray}{0.7} & \textcolor{gray}{0.2}\\ \hline
\texttt{\textbf{BERT-Banter}} & 23.7 & 3.0\\ \hline
\end{tabular}
\caption{\label{table:results} Scores over 20 epochs}
\end{table*}

\subsection{Discussion}

We observed that large values of $scale$ harm the model because large auxiliary rewards tend to veer the agent away from its actual goal.\\

One striking finding in Table \ref{table:results} is the fact that all the Model-Zero variants perform very poorly. This is mostly because the auxiliary reward alone without any intermediate rewards is detrimental, since they are much noisier. While intermediate rewards have direct access to the underlying game engine, our auxiliary rewards have access only to the state representation.\\

The Naive-Bayes based models \texttt{TextBlob} and \texttt{Naive-Bayes} do not improve performance either. This is expected, since naive Bayes uses only word-level features that do not capture the semantic meaning in the state representation. However, as expected, Model-Zero still performs worse because of the problems associated with auxiliary rewards.\\

The BERT-Banter model performs as well as but not better than the Vanilla model. This means that it neither helps or harms the model. We hypothesize that this is due to out-of-dataset generalization. Since the Club-Floyd transcripts are obtained from games that are different from the ones we use, the model likely does not generalize well.

However, the other BERT variants outperform the vanilla agents. Both BERT-Stanford and BERT-Traj aggregate $\approx 9$ points more than the Vanilla model. More importantly, their max scores are $1$ and $2$ more than the Vanilla model. This amounts to completing one more level in the game than the Vanilla model, which directly translates to higher task success. BERT-Stanford's sentiment scores, which are trained on a large corpus, help to achieve a higher score. As expected, BERT-Traj, which is fine-tuned on the trajectories, performs even better because of higher domain similarity between the training data and the text-based games.

\section{Analogous Frameworks}

\subsection{Intrinsically Motivated Reinforcement Learning}

\begin{figure}[ht]
\vskip 0.2in
\begin{center}
\centerline{\includegraphics[width=\columnwidth]{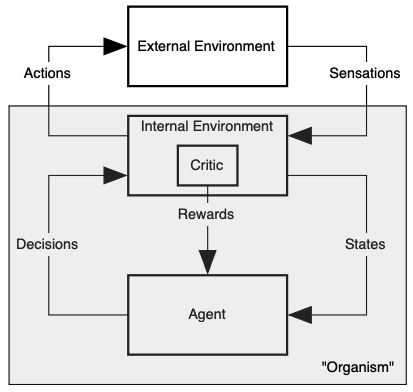}}
\caption{IMRL}
\label{fig:imrl}
\end{center}
\vskip -0.2in
\end{figure}

Our framework is related to the framework of intrinsic motivation, in which the agent rewards itself by analyzing the sensations provided by the environment. In intrinsically motivated reinforcement learning (IMRL), the agent internalizes the reward mechanism, because the same sensations can induce different rewards for different agents (Figure \ref{fig:imrl}). Just as a team's victory can make a person happy or sad depending on the internal reward mechanism of the person, the agent should be able to use the sensations provided by the environment alone to infer its own rewards. Since our model has a sentiment analysis engine that can be internalized into the model, our method can be considered a form of IMRL.

\subsection{Teacher-Student Methods}

An alternative view is to assume that the sentiment analysis engine is a teacher model that provides feedback. While in our case it is a model, we can easily extend it to a human-in-the-loop learning approach like \cite{ross2011reduction}, where the rewards are being provided by humans instead of a model.

\section{Conclusion}

We find that adding auxiliary rewards using sentiment analysis can help improve RL agents' performance in text domains. Our methods take a step in the direction of creating agents that infers rewards by themselves. We expect that these improvements are applicable to similar text-based domains, such as task-oriented dialogue. Given the rapid improvements in NLP methods, we believe that better pre-training and sentiment analysis models will translate to better RL agents in the future.

\section{Acknowledgments}

We thank Shunyu Yao and Rohan Rao for their help with the ClubFloyd transcripts and TextWorld setup. We also thank Karthik Narasimhan and Balaraman Ravindran for helpful comments and feedback.

\bibliographystyle{acl_natbib}
\bibliography{anthology,sarl,refs}

\end{document}